\documentclass[conference]{IEEEtran}
\IEEEoverridecommandlockouts
\usepackage[utf8]{inputenc}
\usepackage[T1]{fontenc}
\usepackage{graphicx}
\usepackage{subcaption}

\usepackage[cmex10]{amsmath}
\usepackage{multirow}
\usepackage{array}
\begin{document}
\IEEEpubid{\makebox[\columnwidth]{}
\hspace{\columnsep}\makebox[\columnwidth]{}}

\title{Light Field Stitching for Extended Synthetic Aperture}
\author{\IEEEauthorblockN{M. Umair Mukati and Bahadir K. Gunturk\thanks{This work is supported by TUBITAK Grant 114E095.}}
\IEEEauthorblockA{Dept. of Electrical and Electronics Engineering, Istanbul Medipol University, Istanbul, Turkey\\
{umairmukati@gmail.com}
{bkgunturk@medipol.edu.tr}
}
}
\maketitle

\begin{abstract}
Through capturing spatial and angular radiance distribution, light field cameras introduce new capabilities that are not possible with conventional cameras. So far in the light field imaging literature, the focus has been on the theory and applications of single light field capture. By combining multiple light fields, it is possible to obtain new capabilities and enhancements, and even exceed physical limitations, such as spatial resolution and aperture size of the imaging device. In this paper, we present an algorithm to register and stitch multiple light fields. We utilize the regularity of the spatial and angular sampling in light field data, and extend some techniques developed for stereo vision systems to light field data. Such an extension is not straightforward for a micro-lens array (MLA) based light field camera due to extremely small baseline and low spatial resolution. By merging multiple light fields captured by an MLA based camera, we obtain larger synthetic aperture, which results in improvements in light field capabilities, such as increased depth estimation range/accuracy and wider perspective shift range.
\end{abstract}

\IEEEpeerreviewmaketitle

\IEEEpubidadjcol

\section{Introduction}

Light field imaging devices capture the amount of light coming from different directions separately, in contrast to the traditional imaging devices, where the directional light information is lost. The idea of measuring
the amount of light from different directions was first implemented by Lippmann \cite{lippmann}, who placed a micro-lens array on a film to record light in
different directions and called this technique ``integral photography''.
Gershun \cite{gershun} worked on the formulation of the distribution of light in
space and used the term ``light field'' for the first time. Adelson and
Bergen \cite{adelson} defined light field as a five dimensional function (3
dimensions for position in space and 2 dimensions for direction). With
the fact that the dimensionality reduces to four in free space where
there is no loss of energy, Levoy and Hanrahan \cite{levoy} and Gortler et
al. \cite{gortler} proposed to analyze light field in a four dimensional parametric
space and paved the way for many applications and theoretical developments today. There are two popular ways of capturing light field. One is to use an array of cameras \cite{levoy,yang} and the other is to use a micro-lens array (MLA) in front of an image sensor \cite{ng2006digital,Lumsdaine}. There are also other light field capture systems, such as coded mask \cite{Veeraraghavan}, lens array \cite{Georgiev}, camera moved on a gantry \cite{Unger}, and kaleidoscope-like optics \cite{Manakov}.

The biggest problem with light field imaging today is low spatial resolution.
There is essentially a trade-off between angular resolution and
spatial resolution; and many light field cameras sacrifice spatial
resolution to gain angular information. For instance, in the first generation
Lytro camera \cite{lytro}, the spatial resolution is less than 0.15 megapixels, while the angular resolution is 11x11. Such a spatial resolution is quite small in today's standards, limiting the use of light field cameras. To address the resolution issue, software based approaches that utilize image restoration techniques have been proposed \cite{Bishop,Cho}.

Once a light field is recorded, images with different camera
parameters can be formed computationally. A regular image can be
formed by adding up all light rays at each pixel. Additionally, aperture, size and shape, focus, point of view, angle of view can be changed; depth estimation can be done; virtual image plane of arbitrary position and orientation can be formed; and geometric aberrations can be corrected.

So far in the light field literature, the focus has been on the processing of single light field capture. Through merging multiple light field data, it is possible to obtain new capabilities and even address some of the fundamental issues of light field cameras, such as limited resolution. In this paper, we present an algorithm to register and stitch of multiple light fields, and generate larger synthetic aperture.

In Section 2, existing approaches on light field stitching are briefly explained. The pre-processing steps before the registration process are presented in Section 3. The proposed registration algorithm is explained in Section 4. The experimental results are provided in Section 5. Conclusions and future work are given in Section 6.

\section{Related work}
Registration of multiple light field captures has recently been addressed in a few publications. In \cite{birklbauer}, a method for creating panoramic light fields is presented. The method is based on projecting two-plane parameterized light fields on a cylindrical coordinate system. The method is limited to rotational motion between light fields; thus, the light field camera must be rotated around its focal point. This requires fixing the camera on a tripod and precise alignment of the rotational center of the tripod with the focal point.

The method presented in \cite{xinqing} is not restricted to rotation around the optical center, and can handle translation as well rotation. The method is based on transforming the light field ray parameters to Pl\"{u}cker coordinates, which results in a projective transformation, named ray-space motion matrix (RSMM), between two light fields. SIFT features are extracted from sub-aperture views to determine the ray correspondences; and the RSMM is estimated from the ray correspondences. It is reported that the method requires large overlap between the light fields to have enough ray correspondences and even with large overlaps rays may not match exactly due to undersampling. This may cause imperfect RSMM estimates, and a graph-cut based refinement step is utilized. One drawback of the method is the high computational cost: The average time to stitch a pair of light fields (captured by a Lytro camera) is about 20 minutes (on a PC with Intel i7 CPU with 64GB memory). Another Pl\"{u}cker coordinate system based approach is presented in \cite{ole}. Ray correspondences are also determined using SIFT features; and the optimization is done based on \cite{li08}.

It should be noted that creating a panoramic light field requires the camera to be rotated around the optical center as in \cite{birklbauer}. When the translation of the camera is allowed, an attempt to create panoramic light field may suffer from ``ghosting artifacts'' due to translation parallax \cite{xinqing}. Because of this fundamental issue, it may be a better idea to generate extended light field aperture instead of attempting to create panoramic view when there is translation of light field camera.

In this paper, we register and merge multiple light fields to obtain a light field with larger synthetic aperture. Different from the previous methods, our registration approach is based on the epipolar geometry of light field data. While epipolar geometry based registration has been studied extensively for structure from motion, the application for light field data is not straightforward when the data is captured with a micro-lens array based camera, such as the Lytro, which has low spatial resolution, low signal-to-noise ratio, and narrow baseline between the sub-aperture images. We show that our approach successfully works with such data.

\section{Light field pre-processing}
\begin{figure}
  \centering
    \includegraphics[width=2.0in]{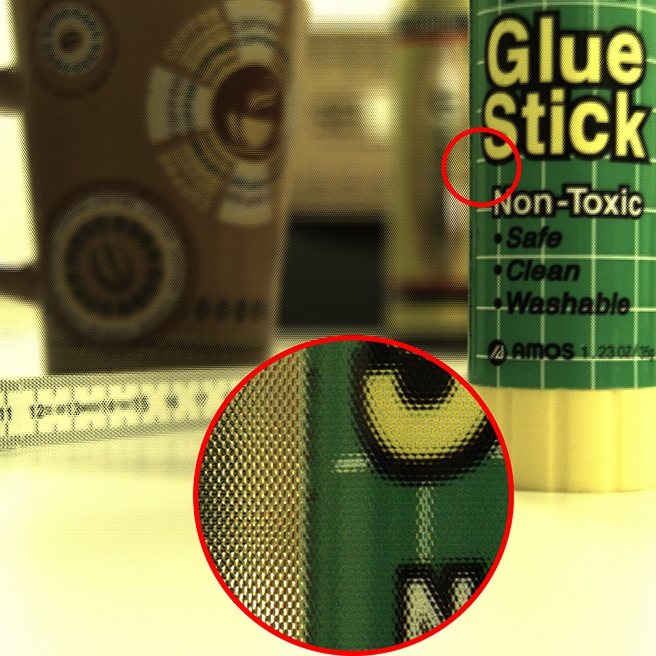}
    \includegraphics[width=2.0in]{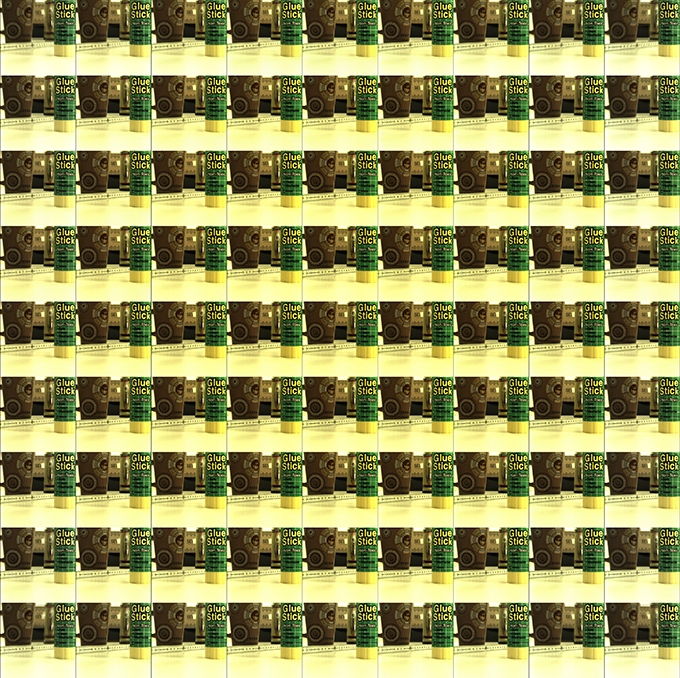}
    \caption{Raw light field data and the decoded sub-aperture (perspective) images.}
	\label{fig_decodedsubaperture}
\end{figure}

We use a first-generation Lytro camera in our experiments. Although the manufacturer does not provide the decoded light field, there are several tools developed to decode light field from raw capture \cite{Cho,sabater,dansereau}. We use the MATLAB toolbox provided in \cite{dansereau} to decode light field from a Lytro raw image capture. From a light field capture, we extract a 9x9 array of sub-aperture images, each with size 380x380 pixels\footnote{While the decoder \cite{dansereau} produces an angular resolution 11x11, we discard the border images as they have poor signal to noise ratio due to severe vignetting.}. In Figure \ref{fig_decodedsubaperture}, a raw light field data and the decoded sub-aperture images are shown.

There are two main pre-processing steps performed on the decoded images before proceeding with the stitching process. The first one is vignetting correction. The intensity of sub-aperture images decreases from middle to side perspectives due to vignetting. To compensate for it, we first apply a Gaussian filter (of size 5x5 and with standard deviation 0.6) to reduce noise in all sub-aperture images, and then estimate and apply the histogram-based photometric mapping \cite{grossberg} to each sub-aperture image to match the colors of the middle perspective image.

\begin{figure}
\centering
    \includegraphics[width=3in]{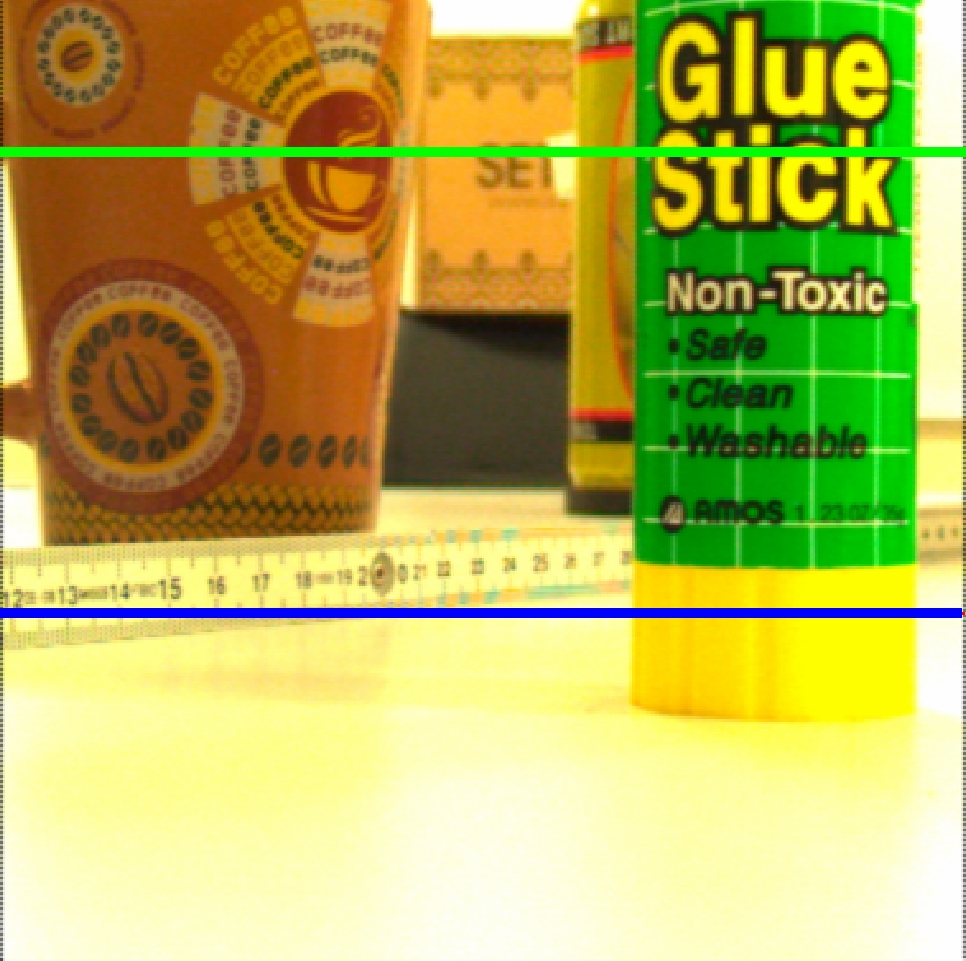}\\
    (a)\\
    \includegraphics[width=3in]{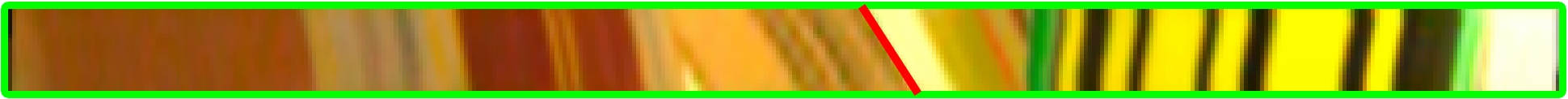}\\
    (b)\\
    \includegraphics[width=3in]{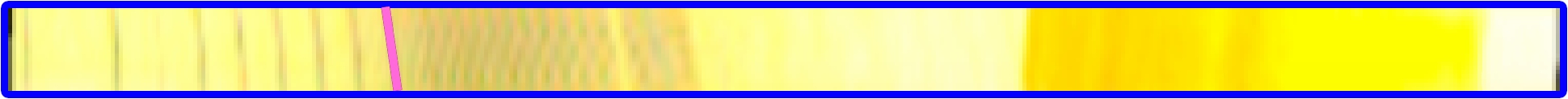}\\
    (c)\\
  \caption{(a) Middle sub-aperture image with two EPI lines marked. (b) EPI for the green line. Largest slope within the EPI is marked with a red line. (c) EPI for the blue line. Largest slope within the EPI is marked with a pink line. The largest slope among all EPIs is selected and used to compensate for the image center shifts.}
  \label{fig_EPIrectification}
\end{figure}

The second pre-processing step is image center correction. As a result of the decoding process, the sub-aperture images might be translated such that the camera array is focused at some mid-range depth. This is clearly seen in the epipolar plane images (EPIs) in Figure \ref{fig_EPIrectification}, where the EPIs include lines with slope larger than 90 degrees (measured from the positive x-axis in the counter-clockwise direction). The largest slope would be 90 degrees if the array were focused at the farthest depth in the scene. Furthermore, it is not guaranteed that the array focuses at the same depth from one light field capture to another. To have the same common reference plane among all light fields, which will be used during the stitching process, we translate all sub-aperture images in a light field to ensure focusing at the farthest depth in the scene. We use the EPI slope based approach \cite{epipolarimages} to estimate the translation amount: The Hough transform \cite{duda1972use} is used to determine all slopes in the EPIs; the largest slope is determined among all EPIs, and each sub-aperture image is translated accordingly. (The process is repeated for horizontal and vertical directions.)

\begin{figure}
\centering
\includegraphics[width=3.5in]{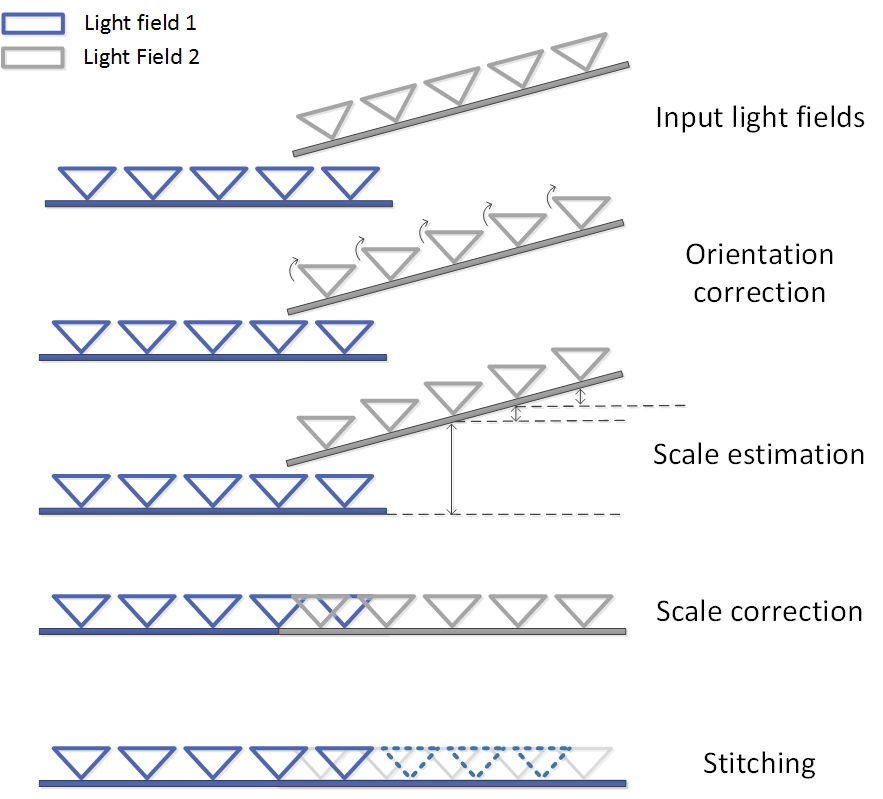}
\caption{Light field rectification and stitching illustrated with virtual cameras capturing sub-aperture images. The first light field is taken as the reference light field; and the second light field is rectified and stitched. The second light field images are rotated to compensate for the orientation difference of the light field cameras, scaled to compensate for the z-axis translations, and finally stitched to the first light field.}
\label{fig_LFrectification}
\end{figure}

\section{Light field registration}

Our light field registration approach consists of rectification and stitching steps. During rectification, all sub-aperture images are compensated for rotation and translation so that they are on the same plane. During stitching, the rectified sub-aperture images are merged into a single light field. We now detail these steps.


\subsection{Rectification of sub-aperture images}

A light field camera can be modeled as an array of virtual cameras, each capturing a sub-aperture (i.e., perspective) image. In case of the Lytro camera, the regularity of the micro-lens array in front of the sensor results in sub-aperture images captured by virtual cameras with regular spacings and identical orientations. In Figure \ref{fig_LFrectification}, we provide an illustration with a virtual camera array as a light field camera, and two light fields captured. (We explain the algorithm for stitching two light fields; the process is repeated for each additional light field.) The sub-aperture images of the second light field are rotated and translated with respect to the first light field sub-aperture images. While the translations differ, the rotation amount between a virtual camera of the first light field and a virtual camera of the second light is identical. First, we correct the orientations of the second light field sub-aperture images. After orientation correction, we correct for the scale to place both light fields onto the same plane.


\subsubsection{Orientation correction}

The orientation difference can be estimated through the fundamental matrix of any sub-aperture image pair from the first and second light fields. We use the middle sub-aperture images of each light field, and estimate the fundamental matrix through feature correspondences as done in traditional stereo imaging systems \cite{hartley}. We extract the Harris corner features \cite{harris1988combined} in the middle sub-aperture image of the first light field and use the Kanade-Lucas-Tomasi (KLT) algorithm \cite{tomasi1991detection} to obtain the correspondences in the middle sub-aperture image of the second light field. The fundamental matrix is then estimated after moving the outliers from the correspondences.

To clarify further, suppose that the corresponding feature coordinates are $(u_i,v_i)$ and $(u'_i,v'_i)$ in the middle sub-aperture image of the first light field and the middle sub-aperture image of the second light field. We then apply the RANSAC technique to remove outliers from the correspondences such that the fundamental matrix equation, $[u_i,v_i,1] F [u'_i,v'_i,1]' = 0$, where $F$ is the fundamental matrix, is satisfied. After the outliers are removed; we estimate the fundamental matrix that minimizes the re-projection error using the gold standard technique \cite{hartley}.

Using the intrinsic matrix $K$, whose parameters (i.e., pixel pitch and focal length) are available in the light field meta data, we calculate the essential matrix $E=K^T F K$. The essential matrix is then decomposed to obtain the rotation matrix \cite{essential}. Specifically, the essential matrix is first decomposed using singular value decomposition (SVD)
\begin{equation}
E = U\Sigma V^{T}
\end{equation}
where $U$ and $V$  are orthonormal matrices and $\Sigma=diag\{\sigma_1,\sigma_2,\sigma_3\}$ is a diagonal matrix, with $\sigma_1$,$\sigma_2$, and $\sigma_3$ being the diagonal elements. For an essential matrix, the first two diagonal elements must be identical and the third element must be equal to zero. To impose this condition, a revised essential matrix is constructed with an updated diagonal matrix $\Sigma=diag\{(\sigma_1+\sigma_2)/2,(\sigma_1+\sigma_2)/2,0\}$, which is optimal in terms of the Frobenius norm \cite{ma}. The new essential matrix is decomposed again using SVD: $E = U\Sigma V^{T}$, and the rotation matrix $R$ is calculated as:
\begin{equation}
R = UWV^{T},
\end{equation}
where $W$ takes two possible versions \cite{ma}:
\begin{equation}
W =
\begin{bmatrix}
0&1&0\\-1&0&0\\0&0&1
\end{bmatrix}
\hspace{0.5cm} \texttt{or} \hspace{0.5cm}
W =
\begin{bmatrix}
0&-1&0\\1&0&0\\0&0&1
\end{bmatrix}.
\end{equation}
Among the two possible solutions for the rotation matrix, only one is physically realizable, which is chosen such that the reconstructed points have positive depths \cite{ma}.

The estimated rotation matrix is then applied to every sub-aperture image of the second light field to correct for the orientation using the homographic transformation $[\alpha u{''}, \alpha v{''}, \alpha]^T =  K R K^{-1} [u', v', 1]^T$, where $(u',v')$ are the pixel coordinates in a sub-aperture image and  $(u{''},v{''})$ are the transformed coordinates.


\subsubsection{Scale estimation and correction}

After the orientation correction, compensation for the z-axis translations (i.e., translations orthogonal to the first light field image plane) within the second light field and between the first and second light fields is required. The effect of these translations is scale change between the images. The scale of each sub-aperture image from the second light field needs to be calculated separately.

\begin{figure}
\centering
\includegraphics[width=3.0in]{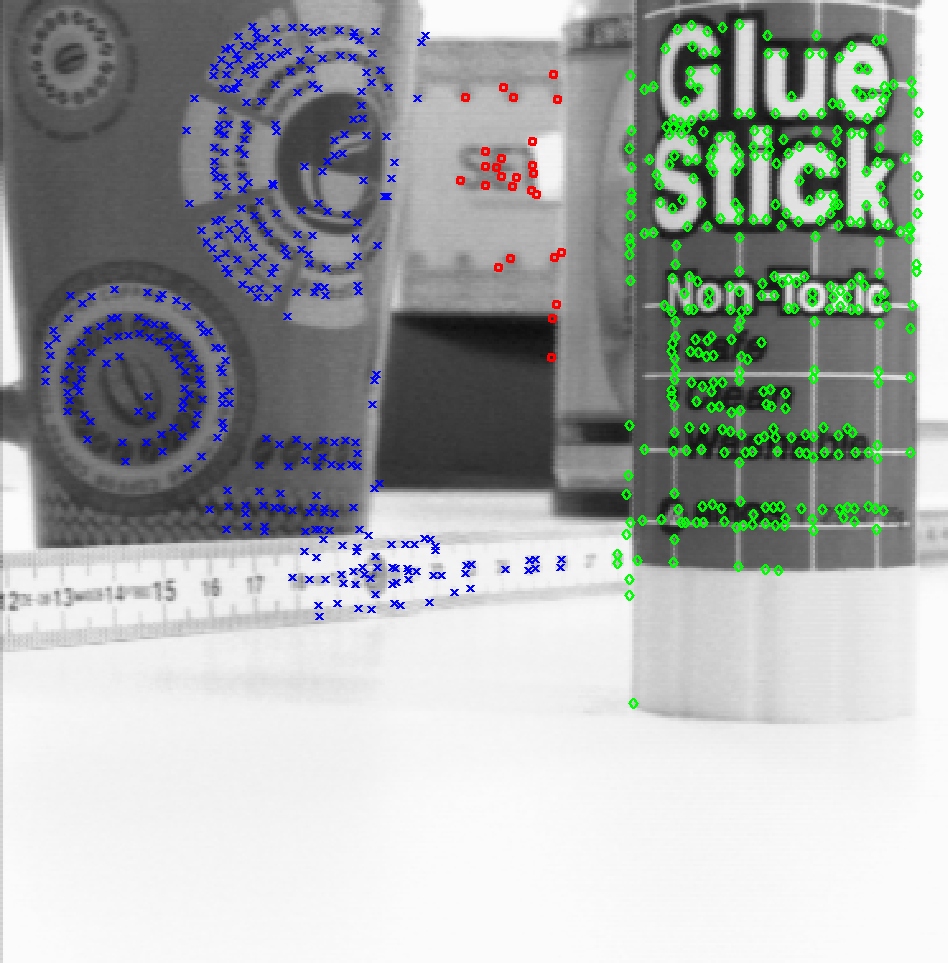}
\caption{Extracted and depth clustered features.}
\label{fig_features}
\end{figure}

{\it Within-light-field scale estimation:} Because the scale is fixed between consecutive pairs of the second light field sub-aperture images, we estimate the scale between every consecutive pair within the second light field and take the geometric mean to have a robust estimate. The scale estimation is again based on feature correspondences. We followed the same procedure (Harris corner detection followed by KLT-based feature tracking) to obtain the feature correspondences. To properly estimate the scale, we should use features from the same depth. The histogram of distances between the correspondences reveal the number of depths available in the scene. We extract the number of depth clusters in our scene according to the Silhoutte's criterion \cite{rousseeuw1987silhouettes} through fitting mixture of Gaussians over the distribution. Features are assigned to a cluster based on their Euclidean distances to the cluster centroids. (The extracted and clustered features from the light field given in Figure \ref{fig_decodedsubaperture} are shown in Figure \ref{fig_features} as an example.)
To estimate the scale, features from any depth cluster can be used; we used the features from the farthest depth cluster. We fit similarity transformation to the feature correspondences between a pair of sub-aperture images to get the scale between the pair.

{\it Between-light-field scale estimation:} The scale between the light fields are estimated by applying the same procedure described above on the middle sub-aperture images of the first and second light fields.

{\it Scale correction:} The estimated within-light-field scales and between-light-field scale are multiplied to obtain the overall scale of each sub-aperture image of the second light field. These scales are then applied to bring all sub-aperture images on the same plane.

\begin{figure}
\centering
\includegraphics[width=3.5in]{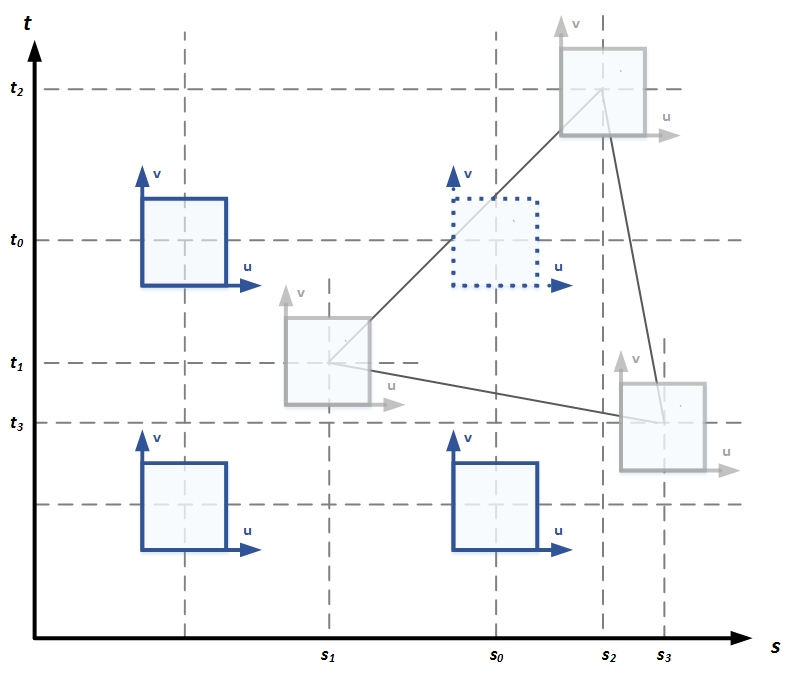}
\caption{Interpolation of sub-aperture images on a regular grid from rectified sub-aperture images.}
\label{fig_intrepRegularGrid}
\end{figure}

\subsection{Light Field Stitching}

The final step is to merge the light fields into a single one. While the sub-aperture images are now all rectified (rotated and scaled), the translation amounts are yet to be determined. We again use a feature correspondence based approach to the determine translations. Using feature correspondences, we first estimate the within-light-field translation amounts between two consecutive sub-aperture pairs. Since the translation amount is fixed between two consecutive pairs, we estimate the translation between every pair and average them to have a robust estimate. We then estimate the translation between the light fields using the middle sub-aperture images. Combining within-light-field and between-light-field translations, we obtain the translations for every sub-aperture image. The translation amounts may not correspond to regular grid locations; in order to obtain a light field on a regular grid, we need to do interpolation. We use the Delaunay triangulation technique for interpolation. As shown in Figure \ref{fig_intrepRegularGrid}, we triangulate the irregular positions of the light fields, and obtain new sub-aperture images at uniform grid positions using pixel-by-pixel weighted sum of neighboring sub-aperture images: Referring to Figure \ref{fig_intrepRegularGrid}, suppose that $(s_0,t_0)$ is the grid position where we have to estimate the sub-aperture image, and $(s_i,t_i)$ with $i=1,2,3$ are the locations where the light field sub-aperture images $I(u,v,s_i,t_i)$ are recorded. If $(s_0,t_0)$ is equal to one of the recorded sample location $(s_i,t_i)$, then the sub-aperture image is directly set to the recorded sub-aperture image at that location. Otherwise, the sub-aperture image $I'(u,v,s_0,t_0)$ is interpolated as a weighted sum of recorded images $I(u,v,s_i,t_i)$, where the weights are inversely proportional to the sample distances:
\begin{equation}
I'(u,v,s_0,t_0) = \frac{ \sum_{i=1}^{3} \left( \frac{1}{\| (s_i,t_i) - (s_0,t_0) \| } \right) I(u,v,s_i,t_i) } { \sum_{i=1}^{3} \left( \frac{1}{\| (s_i,t_i) - (s_0,t_0) \|} \right) }.
\end{equation}

\section{Experimental Results}

In this section, we provide experimental results for two datasets captured with a first generation Lytro camera. We use the light field toolbox of \cite{dansereau} to decode the light fields. All implementations are done in MATLAB, running on an i5 PC with 12 GB RAM. The first dataset consists of 9 light fields where the camera movement is mainly in the horizontal direction. The second dataset includes both horizontal and and vertical movements of the Lytro camera, and includes 10 light fields. The pre-processing time per light field is about 16 seconds, and the rectification time per light field is about 10 seconds. The stitching time depends on the final grid size. The extended light field for the first dataset has a final grid of size 9x24. The extended light field for the second dataset has a final grid of size 26x33. The stitching times are 140 and 300 seconds for the first and second datasets, respectively.

\begin{figure}
\centering
\includegraphics[width=3.5in]{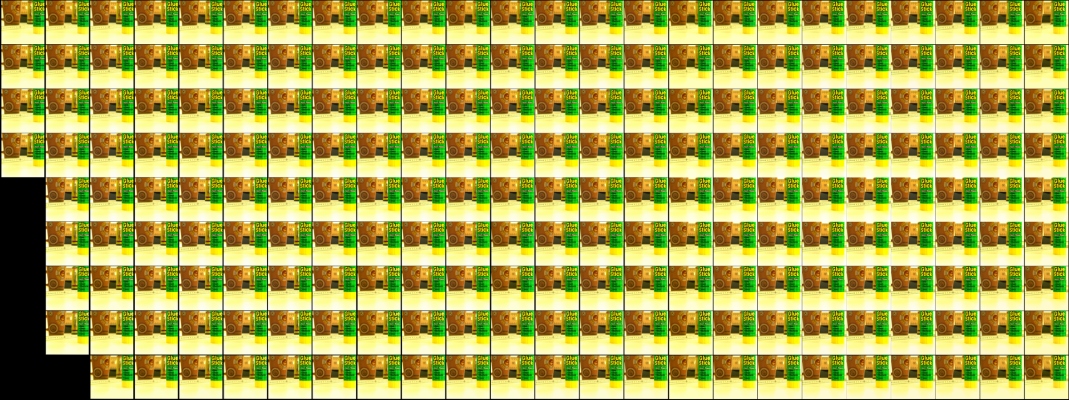}\\
(a)\\
\includegraphics[width=3.5in]{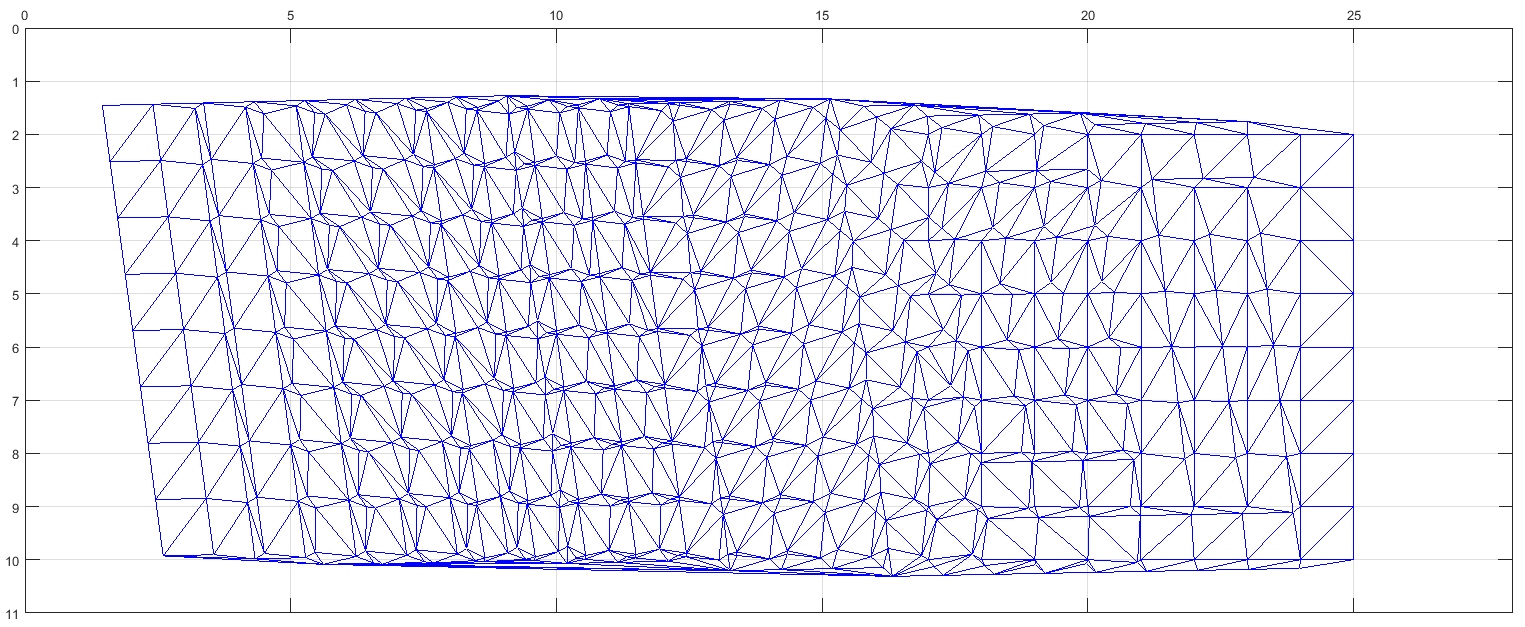}\\
(b)\\
\caption{(a) Final light field obtained by merging of nine light fields (Dataset 1). (b) Estimated sample locations and the resulting Delaunay triangulation.}
\label{fig_extendedLFinterp}
\end{figure}

\begin{figure}
\centering
\includegraphics[width=3.5in]{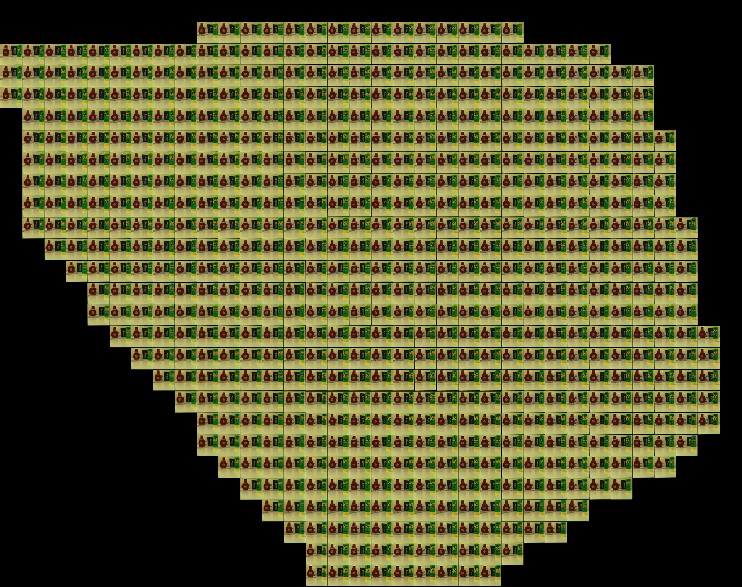}\\
(a)\\
\includegraphics[width=3.5in]{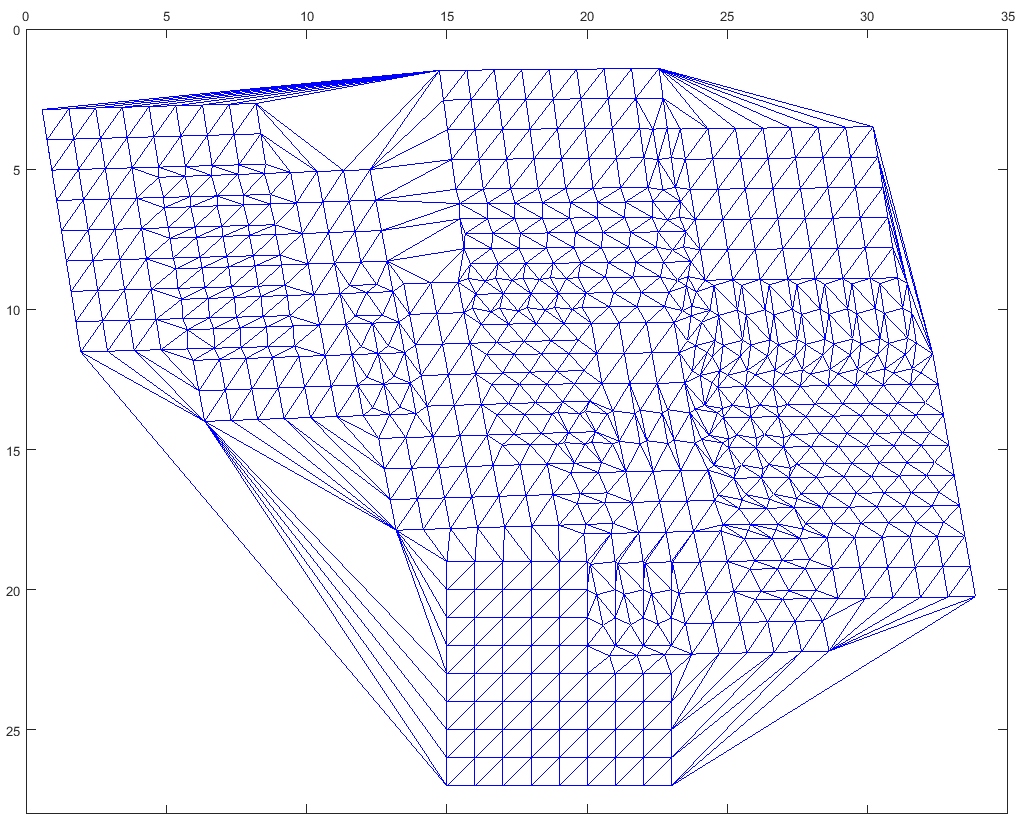}\\
(b)\\
\caption{(a) Final light field obtained by merging ten light fields (Dataset 2). (b) Estimated sample locations and the resulting Delaunay triangulation.}
\label{fig_extendedLFinterp1}
\end{figure}

The extended light field for the first dataset is shown in Figure \ref{fig_extendedLFinterp}(a). The estimated sub-aperture locations and the Delaunay triangulation used to interpolate the missing sub-aperture images are shown in Figure \ref{fig_extendedLFinterp}(b). The extended light field for the second dataset is shown in Figure \ref{fig_extendedLFinterp1}(a); the corresponding sub-aperture locations and Delaunay triangulation used in the interpolation are provided in Figure \ref{fig_extendedLFinterp1}(b).

\begin{figure}
\centering
    \includegraphics[width=3in]{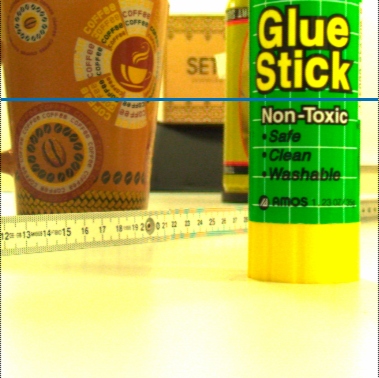}\\
    (a)\\
    \includegraphics[width=3in]{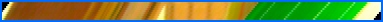}\\
    (b)\\
    \includegraphics[width=3in]{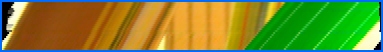}\\
    (c)\\
  \caption{Epipolar plane image extension (Dataset 1). (a) EPI line marked. (b) EPI for the single light field. (c) EPI for the extended light field.}
\label{fig_EPI1}
\end{figure}

\begin{figure}
\centering
    \includegraphics[width=3in]{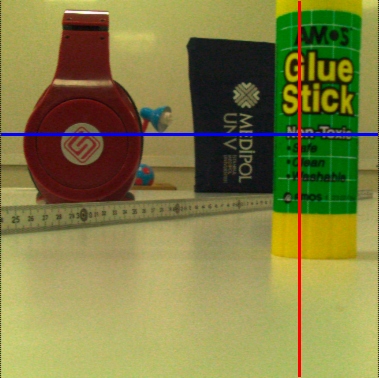}\\
    (a)\\
    \includegraphics[width=3in]{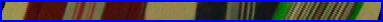}\\
    (b)\\
    \includegraphics[width=3in]{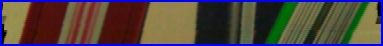}\\
    (c)\\
    \includegraphics[width=3in]{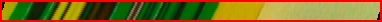}\\
    (d)\\
    \includegraphics[width=3in]{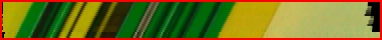}\\
    (e)\\
  \caption{Epipolar plane image extension (Dataset 2). (a) Horizontal and vertical EPI lines marked. (b) EPI for the single light field. (c) EPI for the extended light field. (d) EPI for the single light field. (e) EPI for the extended light field.}
\label{fig_EPI2}
\end{figure}

An EPI example from the first dataset is given in Figure \ref{fig_EPI1}. The EPI demonstrates the extension of the aperture; the straightness of the feature lines in the EPI indicates the correctness of the registration process. In Figure \ref{fig_EPI2}, we show two EPI examples from the second dataset.

\begin{figure}
\centering
  \begin{subfigure}[b]{3.5cm}
    \includegraphics[width=3.4cm]{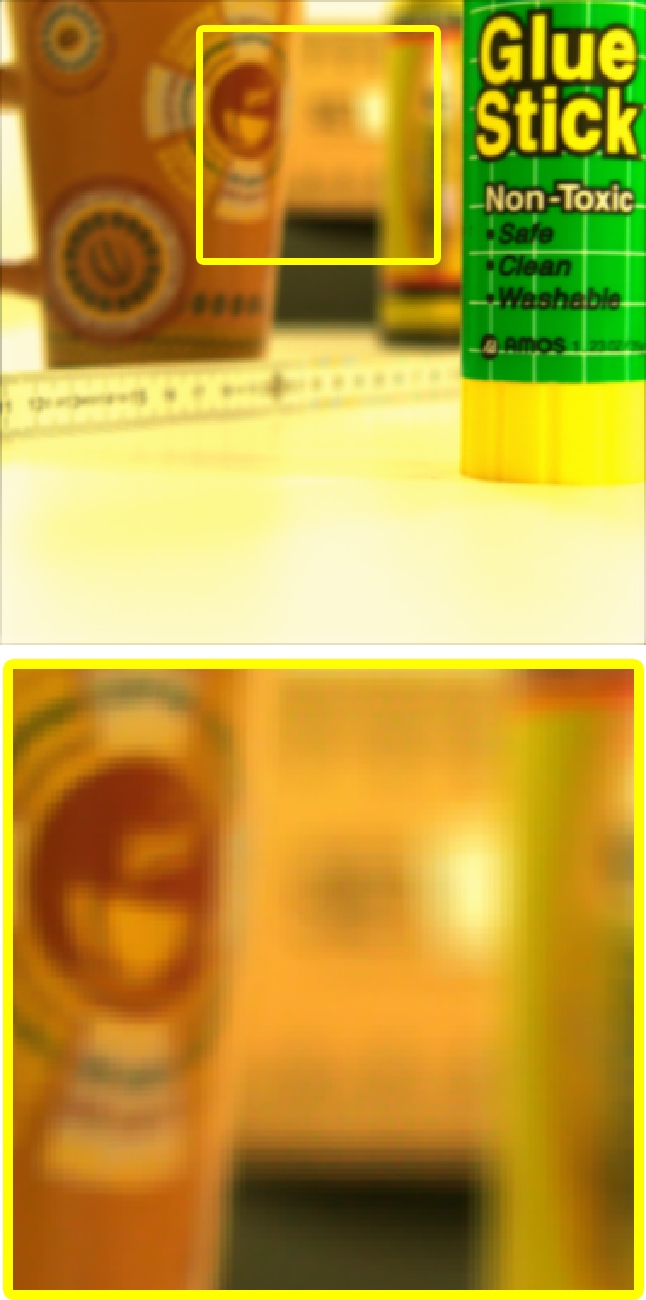}
    \caption{Close focus; single light field.}
  \end{subfigure}
  \begin{subfigure}[b]{3.5cm}
    \includegraphics[width=3.4cm]{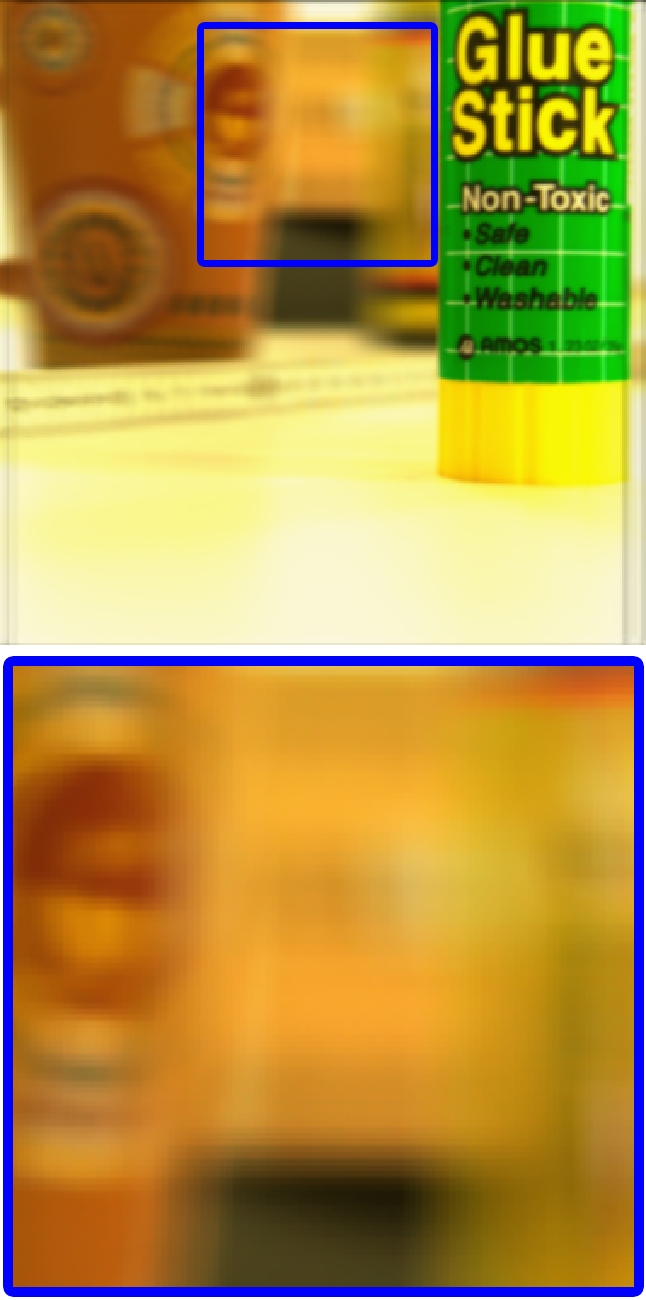}
    \caption{Close focus; extended light field}
  \end{subfigure}
  \begin{subfigure}[b]{3.5cm}
    \includegraphics[width=3.4cm]{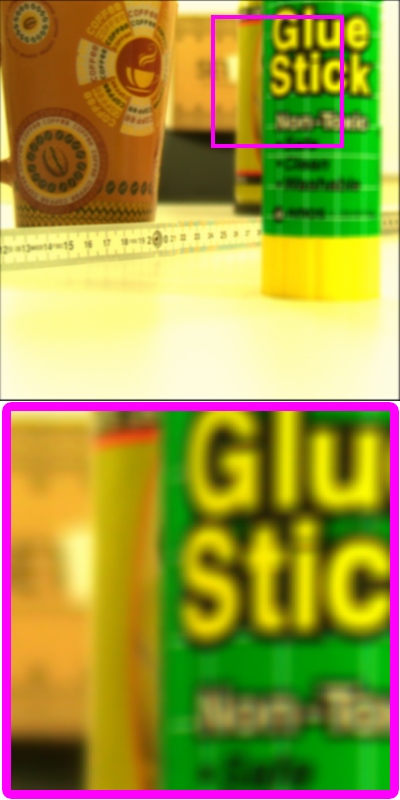}
    \caption{Middle focus; single light field.}
  \end{subfigure}
  \begin{subfigure}[b]{3.5cm}
    \includegraphics[width=3.4cm]{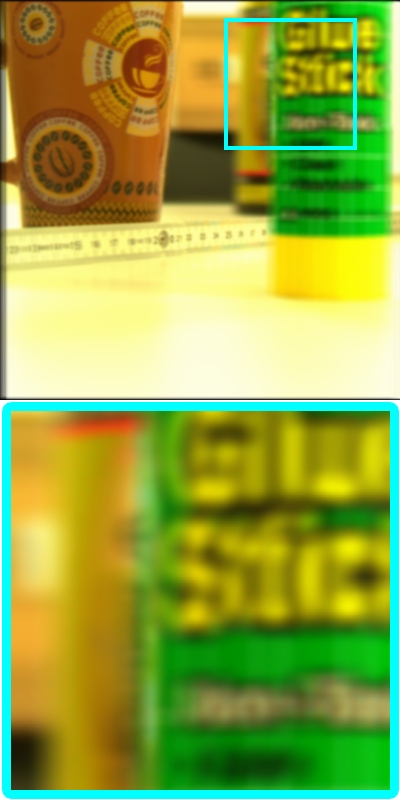}
    \caption{Middle focus; extended light field.}
  \end{subfigure}
  \begin{subfigure}[b]{3.5cm}
    \includegraphics[width=3.4cm]{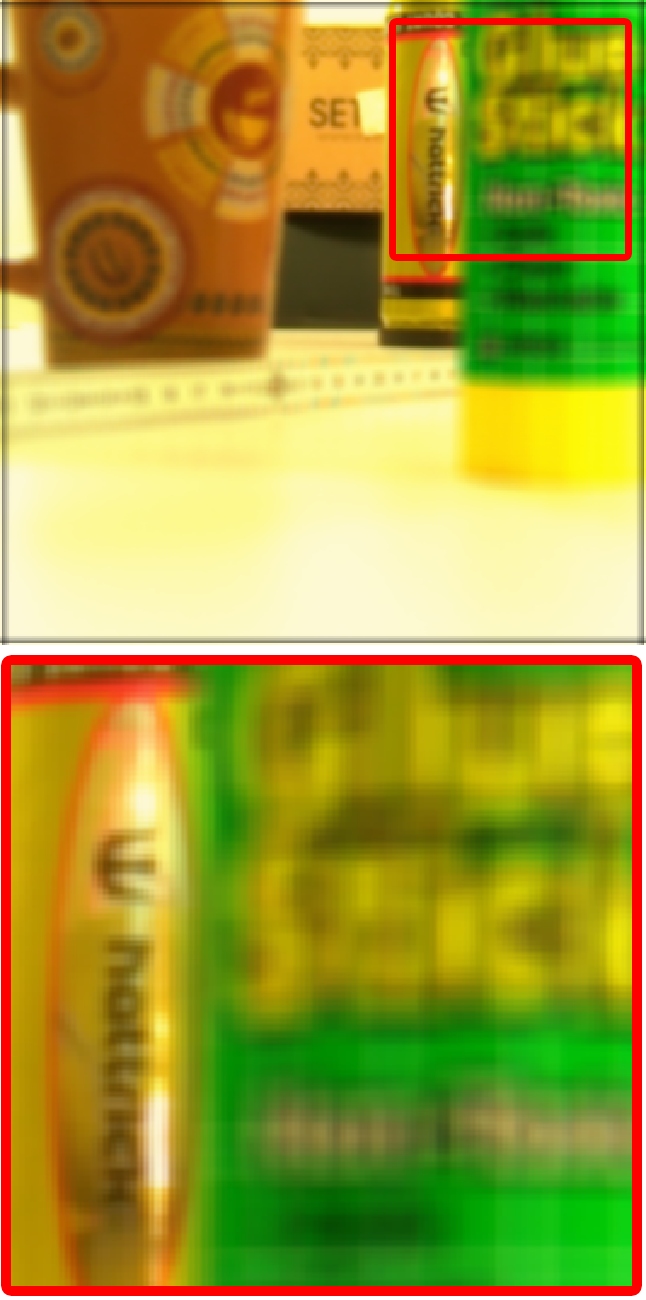}
    \caption{Far focus; single light field.}
  \end{subfigure}
  \begin{subfigure}[b]{3.5cm}
    \includegraphics[width=3.4cm]{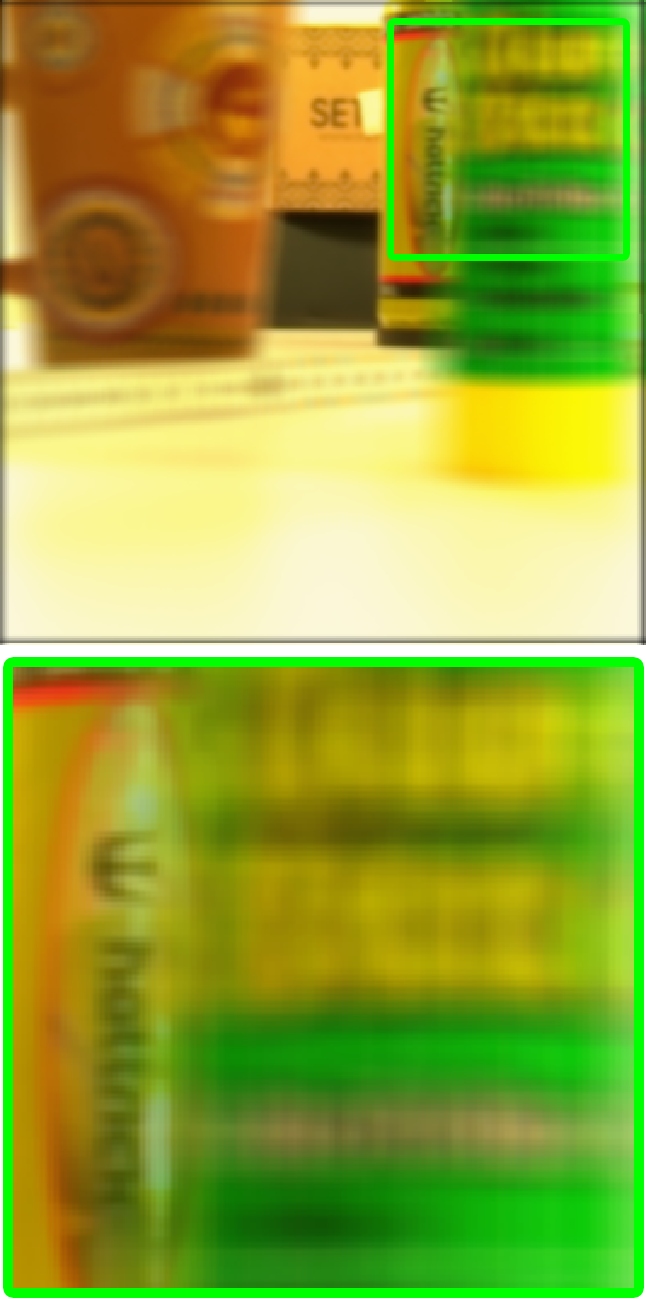}
    \caption{Far focus; extended light field.}
  \end{subfigure}
  \caption{Out-of-focus blurs at different depths are shown (Dataset 1).}
  \label{fig_multipleFocus}
\end{figure}

\begin{figure}
\centering
  \begin{subfigure}[b]{3.5cm}
	\centering
    \includegraphics[width=3.4cm]{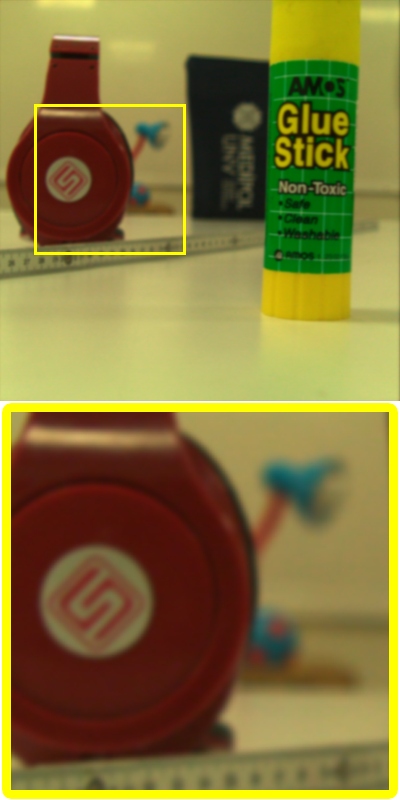}
    \caption{Close focus; single light field.}
  \end{subfigure}
  \begin{subfigure}[b]{3.5cm}
	\centering
    \includegraphics[width=3.4cm]{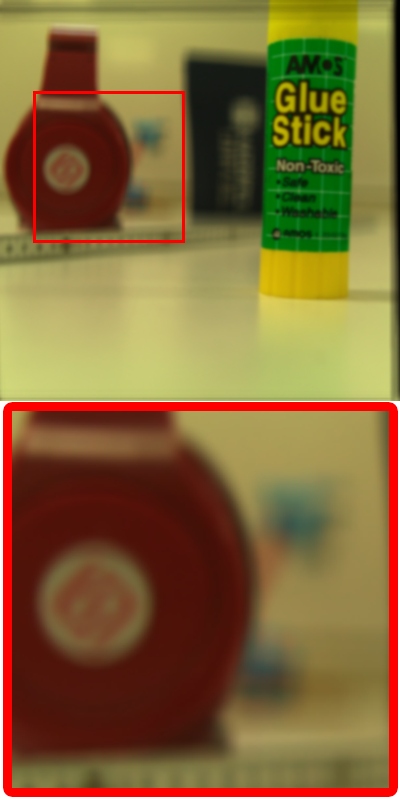}
    \caption{Close focus; extended light field.}
  \end{subfigure}
  \begin{subfigure}[b]{3.5cm}
	\centering
    \includegraphics[width=3.4cm]{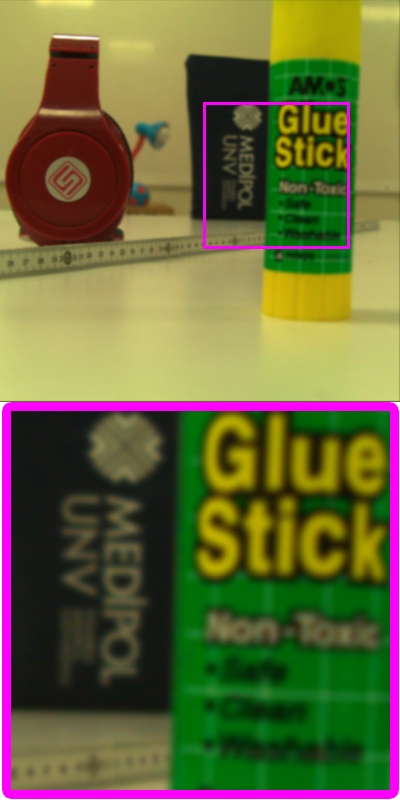}
    \caption{Middle focus; single light field.}
  \end{subfigure}
  \begin{subfigure}[b]{3.5cm}
	\centering
    \includegraphics[width=3.4cm]{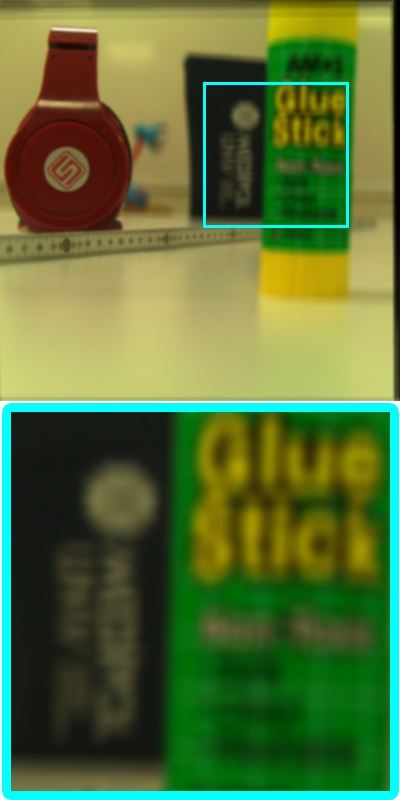}
    \caption{Middle focus; extended light field.}
  \end{subfigure}
  \begin{subfigure}[b]{3.5cm}
	\centering
    \includegraphics[width=3.4cm]{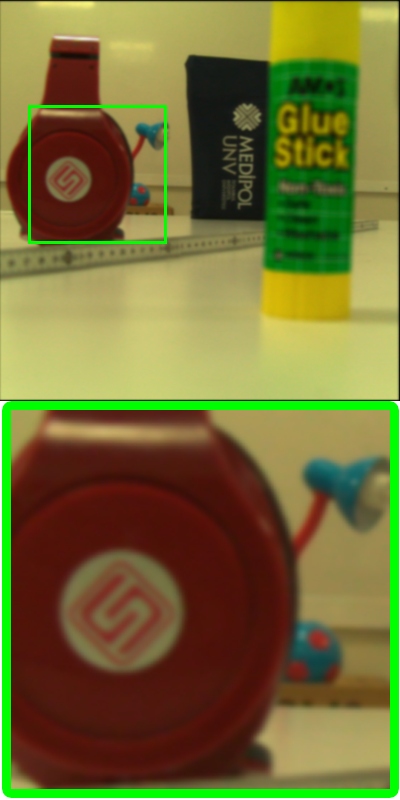}
    \caption{Far focus; single light field.}
  \end{subfigure}
  \begin{subfigure}[b]{3.5cm}
	\centering
    \includegraphics[width=3.4cm]{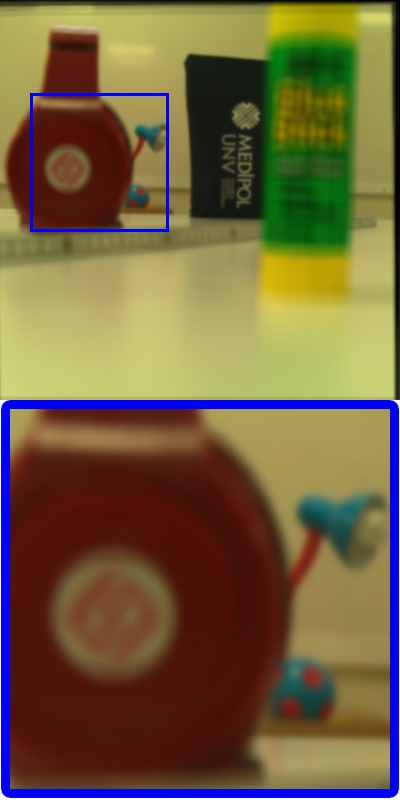}
    \caption{Far focus; extended light field.}
  \end{subfigure}
  \caption{Out-of-focus blurs at different depths are shown (Dataset 2).}
  \label{fig_multipleFocus1}
\end{figure}

{\it Synthetic aperture:} One of the features of light field photography is the ability to digitally change focus after capture. With a larger aperture, the refocusing effect becomes more dramatic as the blur in the out-of-focus regions are larger. In Figures \ref{fig_multipleFocus} and \ref{fig_multipleFocus1}, we focus the light fields at different depths using the shift-and-sum technique \cite{levoy}. The sharpness of the images in the focused regions indicates that the light fields are properly registered. The amount of blur in the out-of-focus regions is larger due to the extended aperture. It can also be noticed that the direction of the blur reflects the extension of the aperture. For example, in Figure \ref{fig_multipleFocus}(d), the blur is more in the horizontal direction, while in Figure \ref{fig_multipleFocus1}(d), the blur is more in the vertical direction.

\begin{figure}
\centering
    \includegraphics[width=2.9in]{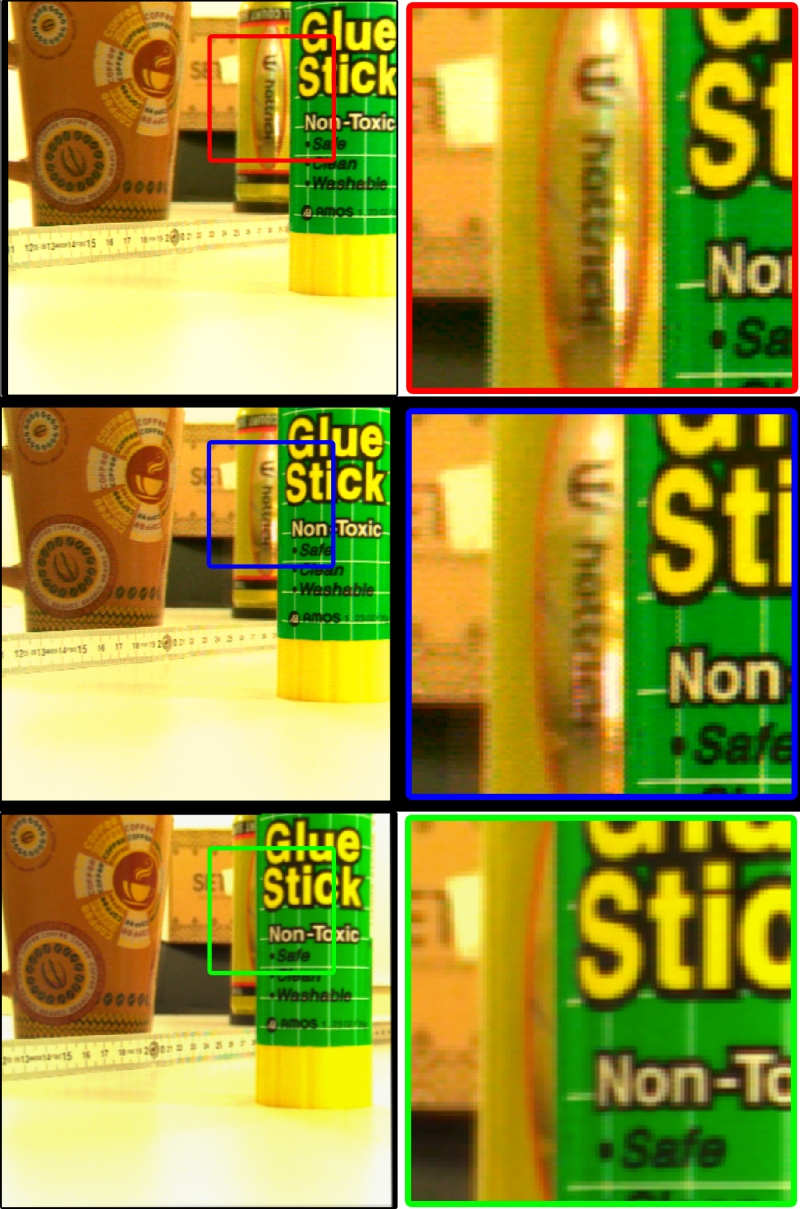}
    \caption{Translation parallax with the single light field and the extended light field. (Top) Leftmost sub-aperture image in the single light field and the extended light field. (Middle) Rightmost sub-aperture image in the single light field. (Bottom) Rightmost sub-aperture image in the extended light field (Dataset 1).}
    \label{fig_transparallax}
\end{figure}

\begin{figure}
\centering
    \includegraphics[width=2.9in]{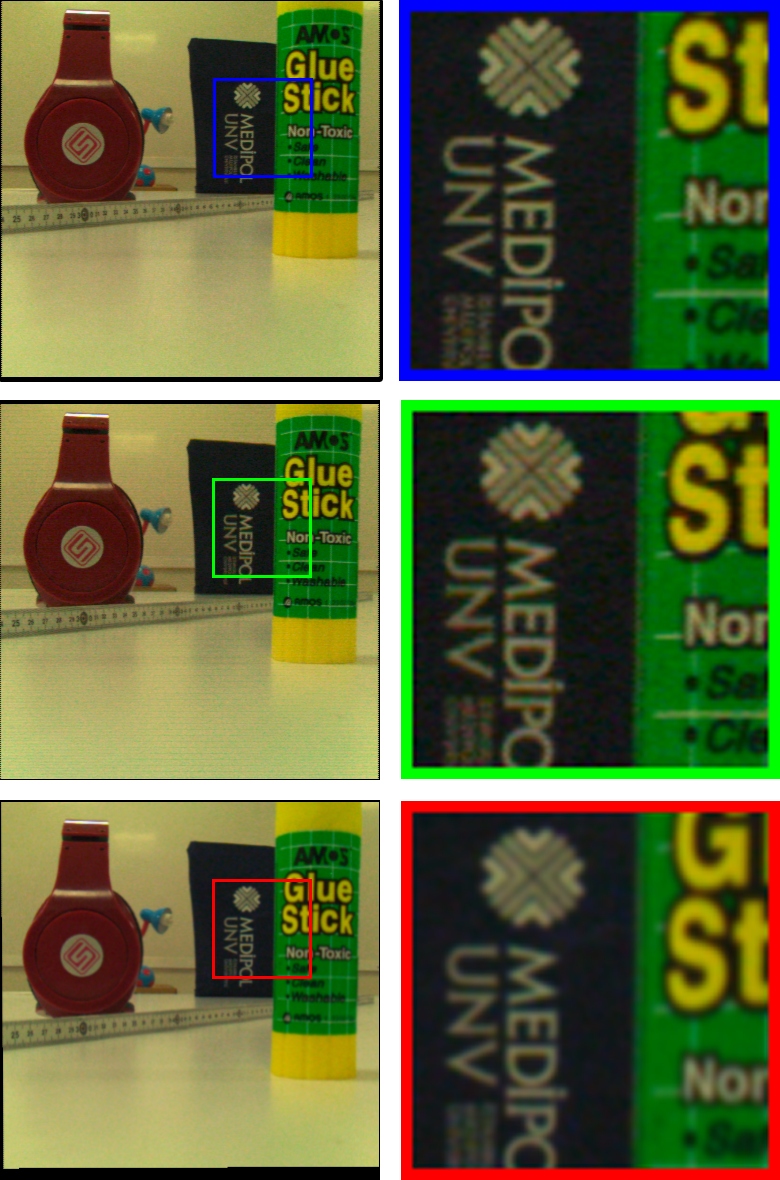}
    \caption{Translation parallax with the single light field and the extended light field. (Top) Bottommost sub-aperture image in the single light field and the extended light field. (Middle) Topmost sub-aperture image in the single light field. (Bottom) Topmost sub-aperture image in the extended light field (Dataset 2).}
    \label{fig_transparallax1}
\end{figure}

{\it Translation parallax:} With the extension of aperture, the baseline between the extreme sub-aperture images of the extended light field is also increased. The effect can be clearly seen by comparing the extreme sub-aperture images of a single light field and extended light field.  In Figure \ref{fig_transparallax}, we show horizontal translation parallax for the single and extended light fields: The top image is the leftmost sub-aperture image in the single light field and the extended light field, the middle image is the rightmost sub-aperture in the single light field, and the bottom image is the rightmost sub-aperture in the extended light field. The increase in translation parallax is visible when these images are compared. Similarly, in Figure \ref{fig_transparallax1}, we compare the vertical translation parallax for single and extended light fields.

\begin{figure}
\centering
  \begin{subfigure}[b]{3.7cm}
    \includegraphics[width=3.7cm]{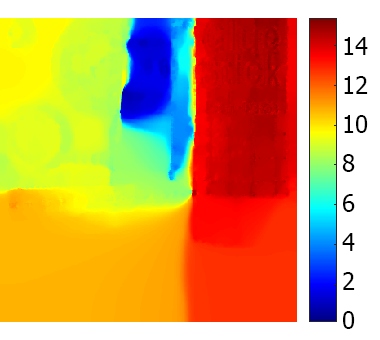}
    \caption{Disparity map with single light field.}
  \end{subfigure}
  \begin{subfigure}[b]{3.7cm}
    \includegraphics[width=3.7cm]{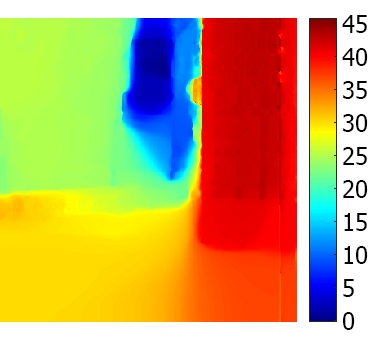}
    \caption{Disparity map with extended light field.}
  \end{subfigure}
  \caption{Disparity map comparison of single and extended light fields for Dataset 1.}
\label{fig_dispmap1}
\end{figure}

{\it Disparity map range:} MLA based light field cameras, such as Lytro, have narrow baseline between the sub-aperture images. This limits the depth map estimation range and accuracy. The relation between baseline and depth estimation accuracy for a stereo system has been studied in \cite{Gallup}, where it is shown that the depth estimation error is inversely proportional with the baseline and increases quadratically with depth. By extending light field aperture, we essentially increase the baseline, which inherently improves both depth estimation range and accuracy. In Figure \ref{fig_dispmap1}, we show the disparity map, obtained by optical flow estimation technique \cite{liu2011sift} between the leftmost and rightmost sub-aperture images, for single and extended light fields. As seen in the figure, the range of the disparity map for the extended light field is (about three times) larger than that of the single light field.

\section{Conclusions}

In this paper, we presented a light field registration algorithm to merge multiple light fields, obtaining extended synthetic aperture. We tested the method with light field data captured by a Lytro camera, which makes the problem more challenging due to its low spatial resolution. One possible extension of the proposed method is increase angular resolution in addition to angular range. This can be done through defining a finer grid for interpolation. Another possible extension is to improve spatial resolution through interpolation in spatial domain in addition to interpolation in angular domain. We believe the proposed registration approach can be in other applications, such as light
field video compression and light field object tracking, as well.



\end{document}